\documentclass[11pt]{article}

\usepackage[preprint]{acl}

\usepackage{fontspec}
\usepackage{latexsym}
\usepackage{microtype}
\usepackage{graphicx}
\usepackage{lineno}
\definecolor{darkblue}{rgb}{0, 0, 0.5}
\hypersetup{colorlinks=true, citecolor=darkblue, linkcolor=darkblue, urlcolor=darkblue}

\usepackage{xeCJK}
\newCJKfontfamily\koreanfont{batang.ttf}[
  AutoFakeBold=true]
\newcommand{\kr}[1]{{\koreanfont #1}}
\newfontfamily\tamilfont{HindMadurai-Regular.otf}[
  Script=Tamil,
  BoldFont=HindMadurai-Bold.otf]
\newcommand{\ta}[1]{{\tamilfont #1}}

\usepackage{booktabs}

\title{When does text encoding matter for language modeling?}

\author{Eugene Jang\textsuperscript{1}, Catherine Arnett\textsuperscript{2}, Terra Blevins\textsuperscript{1} \\
  \textsuperscript{1}Khoury College of Computer Sciences, Northeastern University \\
  \textsuperscript{2}EleutherAI \\
  \texttt{\{jang.eu, t.blevins\}@northeastern.edu} \hspace{0.4cm} \texttt{catherine@eleuther.ai} \\
  }

\begin{document}
\maketitle

\begin{abstract}
UTF-8's role as the \textit{de facto} encoding standard in language models introduces cross-lingual inequities due to its variable-length design.
However, the effect that this encoding choice has on model performance remains largely unexamined.
In this work, we investigate how text encoding impacts the performance of byte-level language models, comparing UTF-8 with a wide range of other encoding strategies, including other Unicode-based approaches, language-specific encodings, and our own customized encodings.
To facilitate our analysis, we propose \textsc{Aether}: a hybrid encoding scheme that uses script-specific encodings for our target setting while maintaining full Unicode coverage through UTF-8 fallbacks, using this framework to model four orthographically diverse languages.
Surprisingly, we find that while languages with information-dense characters (Chinese, Korean) significantly benefit from compressive encodings, languages whose characters carry less information individually (Georgian, Tamil) are largely insensitive to encoding choice. 
In light of these findings, we argue that text encoding should be a language-specific design decision, rather than a universal default.\footnote{Code and models available at: \url{https://github.com/genesith/AetherModels}}

\end{abstract}
\section{Introduction}
Subword tokenization (i.e. BPE; \citealp{gagebpe,sennrich-etal-2016-bpe}) has long been the dominant approach to tokenization \citep{mielke2021between}, but there is growing interest in byte-level tokenization (\citealp[]{tay2022charformer,xue-etal-2022-byt5,wang2024mambabyte}, \textit{inter alia}) as a way to overcome the shortcomings of subword tokenization \citep{bostrom2020bytepairencodingsuboptimal,beinborn-pinter-2023-analyzing, wang2025tokenizationmattersdegradinglarge,chai-etal-2024-tokenization}.
While these methods aim to reduce inductive biases carried by subword tokens, the inherent biases of the bytes used as the bases for language models instead are often overlooked.
The \textit{de facto} standard for converting text into bytes in modern computing applications is UTF-8, a variable length encoding scheme capable of expressing all Unicode codepoints using 1$\sim$4 bytes; due to its ubiquity, it is also the primary method for converting text into language modeling inputs. However, UTF-8 introduces inefficiencies~\citep{moon2025bitlevelbpebyteboundary, kim2026bbpe16utf16basedbytelevelbytepair} and unfairness~\citep{Petrov-unfairness, arnett-etal-2024-bit}, leading some to argue that it is suboptimal for NLP applications~\cite{van-der-goot-2026-bytes}.

\begin{figure}[t!]
    \centering
    \includegraphics[width=\linewidth]{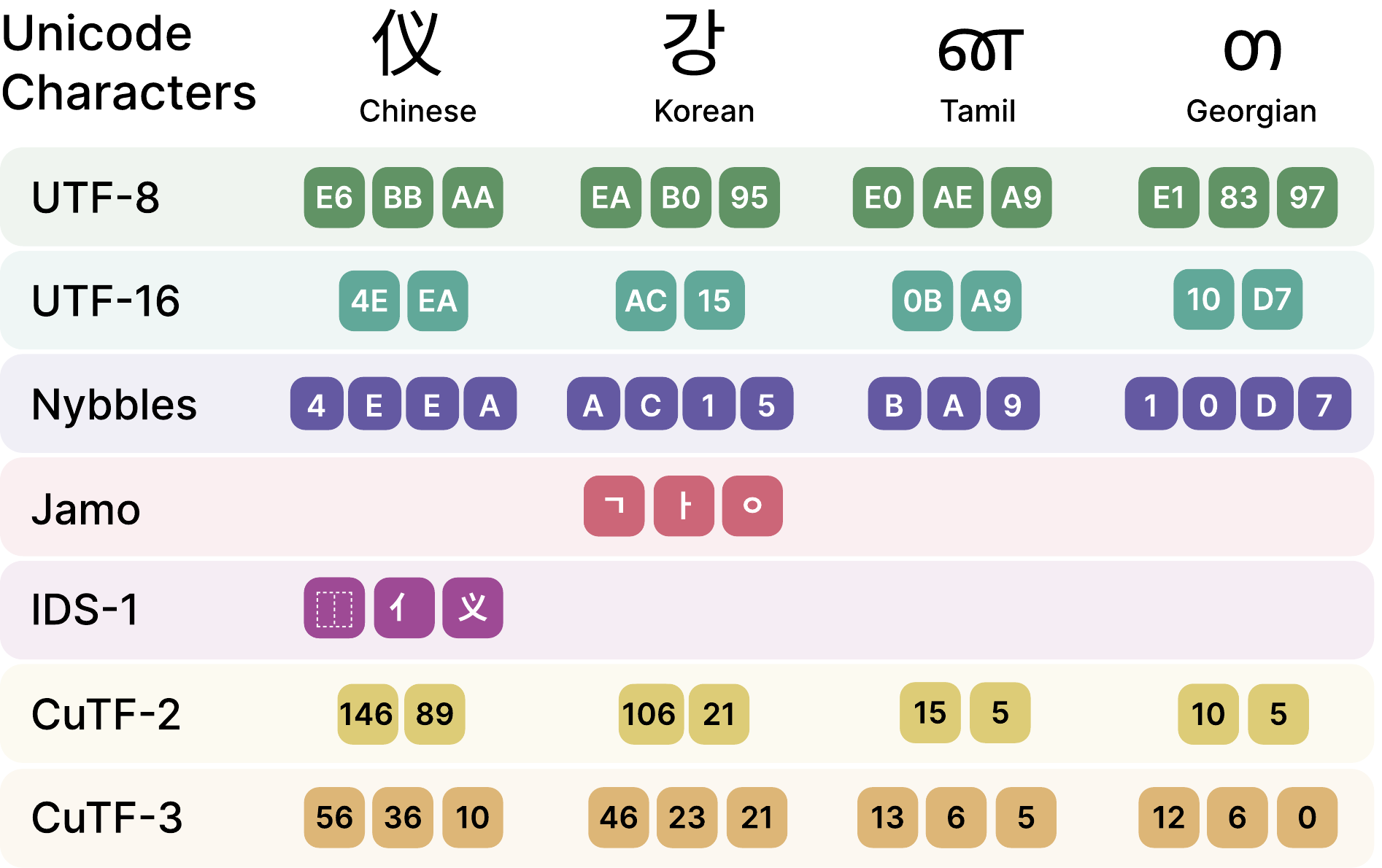}
    \caption{Examples of alternative text encodings, including Unicode-based encodings (UTF-8, UTF-16, Nybbles), language-specific encodings (Jamo, IDS-1), and our new custom encodings (CuTF-2, CuTF-3).
    }
    \label{fig:mainfig}
\end{figure}



Recent work, such as MYTE~\citep{limisiewicz-etal-2024-myte} and SCRIPT~\citep{land2025bpestaysscriptstructured}, propose new encoding strategies designed to address fairness between languages.
However, these studies are limited to a single alternative encoding designed to improve the UTF-8 baseline on a specific dimension, leaving many aspects of how encoding choice affects language modeling unexplored.
In this work, we present a systematic comparison of encoding methods for byte-level language modeling, based on a wide range (16) of encoding methods spanning Unicode-based encodings, language-specific encodings, and our own proposed custom encodings (Figure~\ref{fig:mainfig}).
To enable controlled comparisons within a target script, we propose \textsc{Aether} tokenizers (Alternative Encoding Tokenizers for Hybrid Encoding Representations), which apply script-specific encodings when relevant while applying UTF-8 as a back-off encoding for other text.
Using Aether tokenizers, we train ``byte"-level models on four orthographically distinct languages (Chinese, Korean, Georgian, Tamil) with 38 unique language-encoding pairs.

Our experiments reveal two distinct patterns. For Chinese and Korean, encoding choice has a clear impact on performance, with more compressive encodings yielding significantly better results. In contrast, all encodings perform similarly for Georgian and Tamil, even when their compression rates differ by up to ${\sim}7\times$ and ${\sim}8\times$, respectively. We hypothesize that these divergent results stem from differences in the \textit{orthographic granularity} of the underlying scripts. Specifically, when characters individually encode meaningful information (as in Chinese and Korean), compression appears to heavily impact performance, while characters that carry little information in isolation (like Tamil and Georgian) are largely insensitive to encoding selection.
Given these script-specific trends, we argue that text encoding should be treated as a deliberate design choice that accounts for these properties rather than accepted as a universal default.


\section{Related Work}

\paragraph{Character/Byte-Level Modeling}
The drawbacks of subword tokenizers are often cited as motivations to use atomic units of text as tokens, in the form of characters~\citep{clark-etal-2022-canine, gao-etal-2020-character, lee-etal-2017-fully} or bytes~\citep{xue-etal-2022-byt5,neitemeier2025hierarchical,hwang2025dynamicchunkingendtoendhierarchical}.\footnote{Characters and bytes are often used interchangeably in this context. Several works use ``character-level" in writing while implementing on the byte-level~\citep{al2019character, choe2019bridginggaptokenizerfreelanguage, tay2022charformer}.}
However, these approaches have seen limited adoption, which \citet{libovicky-etal-2022-dont} attribute to their increased computational cost and underperformance to subword counterparts.
On the other hand, \citet{edman-etal-2024-character} identify benefits of byte-level models in low-resource MT scenarios, and argue lower efficiency should be seen as a tradeoff.
\citet{minixhofer2026bolmobyteifyinggenerationlanguage} note that such models are advantageous in domains that benefit from character understanding, such as code and biological sequences.
While byte-level models have some promising advantages, the literature typically defaults to the UTF-8 encoding.


\paragraph{Byte Premiums in UTF-8}
Past work discuss how tokenization length can vary greatly between languages when conveying the same information~\citep{ahia-etal-2023-languages, Petrov-unfairness}.
\citet{arnett-etal-2024-bit} identify how UTF-8 variable byte lengths can systematically distort text dataset sizes.
UTF-8 uses variable byte-lengths to encode different scripts, with higher Unicode code points requiring more bytes.
Basic Latin characters take up early code points represented by single bytes, while other language scripts are encoded with multiple bytes per character.
A script's encoding efficiency therefore depends on its position within Unicode, as earlier code points are encoded as two bytes (e.g. Cyrillic, U+0400--U+04FF) while later code points require three bytes (e.g. Tamil, U+0B80--U+0BFF) or more.
Furthermore, two-byte UTF-8 sequences can represent only up to 1,920 code points, which is insufficient for large script blocks, such as CJK Unified Ideographs (20,992 characters) or Hangul Syllables (11,172 characters).
In this work, we experiment with Chinese, Korean, Tamil, and Georgian, all of whose scripts fall within the three-byte encoding range.

\paragraph{Alternatives to UTF-8}
Previous works explore different encoding methods, either by creating new indexing strategies or modifying UTF-8.
\citet{limisiewicz-etal-2024-myte} propose MYTE encoding, which converts redundant bytes in the UTF-8 bytespace to encode morphological units.
\citet{land2025bpestaysscriptstructured} propose SCRIPT encoding, where all Unicode characters are encoded in two-part representations of block-indexing pairs.
Few works specifically target the large character inventories in the CJK Unified Ideographs and Hangul Syllables blocks.
\citet{kim2026bbpe16utf16basedbytelevelbytepair} suggest representing these code points with UTF-16 encoding, which results in two bytes per character.
Similarly, \citet{moon2025bitlevelbpebyteboundary} compress such code points to two 9-bit units by deduplicating redundancies and redistributing the indexing bits of their UTF-8 byte representations.
Other works leverage properties specific to a target script, such as decomposability or glyph information~\citep{stratos-2017-sub, moon-okazaki-2020-jamo,si-etal-2023-sub}.
We build on these approaches to systematically evaluate many encoding schemes on language model performance.
\section{Alternative Encodings}
\label{sec:alt-encodings}

\begin{table*}[t]
\centering
\small
\resizebox{\textwidth}{!}{%
\begin{tabular}{lcrrrrrrrr}
\toprule
        &        & \multicolumn{2}{c}{Korean (\texttt{KO})} & \multicolumn{2}{c}{Chinese (\texttt{ZH})}& \multicolumn{2}{c}{Tamil (\texttt{TA})} & \multicolumn{2}{c}{Georgian (\texttt{KA})} \\
\cmidrule(lr){3-4} \cmidrule(lr){5-6} \cmidrule(lr){7-8} \cmidrule(lr){9-10}
Encoding & Length & Vocab & Steps/Ep. & Vocab & Steps/Ep. & Vocab & Steps/Ep. & Vocab & Steps/Ep.\\
\midrule
\multicolumn{6}{l}{\textit{Unicode-based methods}} \\
Character   & 1      & 11,172& 23,804	&	20,902	& 23,756 & 128	& 23,756	&	96	& 23,585\\
UTF-16    & 2      & 256	&39,202&		256&	40,768&		129&	41,439&		97&	42,749 \\
UTF-8        & 3      & 68    & 54,600 & 70    & 57,781
& 65    & 60,368	 & 65 &	61,913\\
Nybbles     & 3/4      & 16    & 69,998 & 16    & 74,793 
& 16    & 79,296 & 16    & 81,077 \\
Crumbs & 6/8      & 4   &131,589	&	4	&142,842	&	4	&117,151	&	4	&157,733 \\

\addlinespace
\multicolumn{6}{l}{\textit{Language-specific methods}} \\
CP949       & 2      & 178   & 39,202 & ---   & --- & ---   & --- & ---   & ---   \\
Johab       & 2      & 177   & 39,202 & ---   & --- & ---   & --- & ---   & ---   \\
Jamo-varlen & 2--3   & 67    & 46,472 & ---   & --- & ---   & --- & ---   & ---   \\
Jamo        & 3      & 68    & 54,600 & ---   & --- & ---   & --- & ---   & ---   \\
GBK         & 2      & ---   & ---   & 190   & 40,768 & ---   & --- & ---   & --- \\
IDS-1       & 1--10  & ---   & ---   & 3007  & 33,766 & ---   & --- & ---   & --- \\
IDS-2       & 1--17  & ---   & ---   & 1138  & 50,030 & ---   & --- & ---   & --- \\
TamilGraphemes       & <1  & ---   & ---   & ---  & ---  & 314   & 14,789 & ---   & --- \\
\addlinespace
\multicolumn{6}{l}{\textit{Custom UTFs}} \\
CuTF-2 & 2      & 212   & 39,202 & 290   & 40,768 
& 23   & 41,439 & 20   & 42,749 \\
CuTF-3 & 3      & 68    & 54,600 & 83    & 57,781 
& 16    & 60,368 & 14    & 61,913 \\
CuTF-4 & 4      & 42    & 70,004 & 49    & 74,793 & ---   & --- & ---   & --- \\

\bottomrule
\end{tabular}
}

\caption{Encoding methods tested in our experiments for Korean (\texttt{KO}), Chinese (\texttt{ZH}), Tamil (\texttt{TA}), and Georgian (\texttt{KA}). Vocab size refers to the unique values in encoding inventory used to represent the target script.}
\label{tab:encodings}
\end{table*}

In this section, we describe the encoding methods tested in our experiments (Table~\ref{tab:encodings}).

We consider an encoding to be any mapping of natural language text into an atomic basis for modeling (potentially ranging from bit-based representations to multi-character units).
Critically, we do not restrict ourselves to byte- or bit-aligned units; an encoding may use any indexing unit and vocabulary size we define. 
We group the considered encodings as follows: Unicode-based encodings (§\ref{sec:enc-unicode}); language-specific encodings designed to represent a single script (§\ref{sec:enc-language}); and CuTF, our customized indexing scheme that efficiently encodes a script at a fixed length (§\ref{sec:enc-cutf}).
Examples of encodings are illustrated in Figure~\ref{fig:mainfig}.

\subsection{Unicode-based Encodings}
\label{sec:enc-unicode}
Unicode is a character encoding standard that unifies previously separate character sets by assigning each character a unique code point.
The code points are converted into code units through UTF (Unicode Transformation Format) encodings: UTF-8 uses 8-bit bytes as code units while UTF-16 uses 16-bit byte-pairs as code units.

\paragraph{UTF-8 Encoding}
While UTF-8 has 256 total byte values in its inventory, only a subset of these values are used to encode multi-byte characters.
Each UTF-8 multi-byte character encoding is composed of two types of bytes: the starting byte and the continuation bytes.
Starting bytes use part of their beginning bits to indicate the number of continuation bytes that follow while the remaining bits are used for indexing.
For instance, in the byte 0xEC (\texttt{11101100}) the 4-bit prefix (\texttt{1110}) signals that it begins a three-byte sequence, while the remaining 4 bits (\texttt{1100}) carry indexing information.
Continuation bytes have a fixed 2-bit prefix (\texttt{10}), and use the remaining 6 bits to carry indexing information.
Consequently, only 16 of the 24 bits in a three-byte UTF-8 encoding of a code point contain indexing information.

Each character in scripts in the 3-byte code point range can be expressed in the form of triple $(s, c_1, c_2) \in S_\ell \times C \times C$, where $S_{\ell}$ is the set of possible starting bytes for script $\ell$ ($|S_{Zh}|=6, |S_{Ko}|=4, |S_{Ta}|=1, |S_{Ka}|=1$) and $C$ is the continuation bytes ($|C|=64$).
Continuation bytes $C$ are shared for all multi-byte characters across scripts.

\paragraph{UTF-16 Variants} The larger code unit of 16-bits allows for more compact indexing compared to its 8-bit counterpart for certain code points.
Code points that would require three UTF-8 bytes can be instead represented as a single 16-bit unit (recall that code points in this range only use 16 bits for indexing).
The 16-bit unit can then be interpreted as a two-byte sequence~\cite{kim2026bbpe16utf16basedbytelevelbytepair}.
For our experiments, \textbf{UTF-16} encodings will refer to the big-endian bytes obtained by splitting UTF-16 encodings in half.
We argue that splitting the 16 indexing bits to finer-grained units can be iteratively extended upon.
We propose two such encodings:
\textbf{Nybbles} splits a 16-bit sequence into four nybbles (4-bit units) and \textbf{Crumbs} splits the 16-bit sequence into eight crumbs (2-bit units)\footnote{For Tamil script, we encode characters as 3 nybbles or 6 crumbs, since the first 4-bits are always zero.}.
Finer-grained units reduce the base vocabulary size (from 256 symbols for bytes to 16 for nybbles and 4 for crumbs) at the cost of longer sequences (from 2 to 4 to 8 units per code unit).

\subsection{Language-specific Encodings}
We examine four languages that use different writing systems: Chinese (logography), Korean (syllabary)\footnote{Korean Hangul has been described as an alphabetic syllabary~\cite{Pae01012011}, and has characteristics of an alphabet, syllabary, and logography~\cite{Taylor1980}. We consider it to be closest to a syllabary in the context of text processing, since it is most commonly represented as syllable units in Unicode.}, Tamil (abugida), and Georgian (alphabet).
For each language, we source various representational methods, including legacy encodings and orthography-aware representations.

\label{sec:enc-language}
\subsubsection{Legacy Encodings}
Prior to the wide adoption of Unicode, text encodings were designed to compactly represent the characters used in a target language, but they did not cover all other scripts.
When languages have large character inventories, such as Chinese and Korean, language-specific encodings like \textbf{GBK} for Chinese ideographs and \textbf{CP949} and \textbf{Johab} for Korean Hangul use two bytes to represent characters.
When character inventories are sufficiently small, language-specific encodings can assign a single byte to each character, as GEOSTD8 and TSCII do for the Georgian and Tamil scripts, respectively.
We note that byte-level models using such single-byte encodings are equivalent to training on character-level granularity.
Therefore, rather than testing these encodings directly, we evaluate the general strategy of \textbf{Character}-level encoding, applying it to all four languages regardless of character inventory size.


\begin{figure*}[t!]
    \centering
    \includegraphics[width=0.8\linewidth]{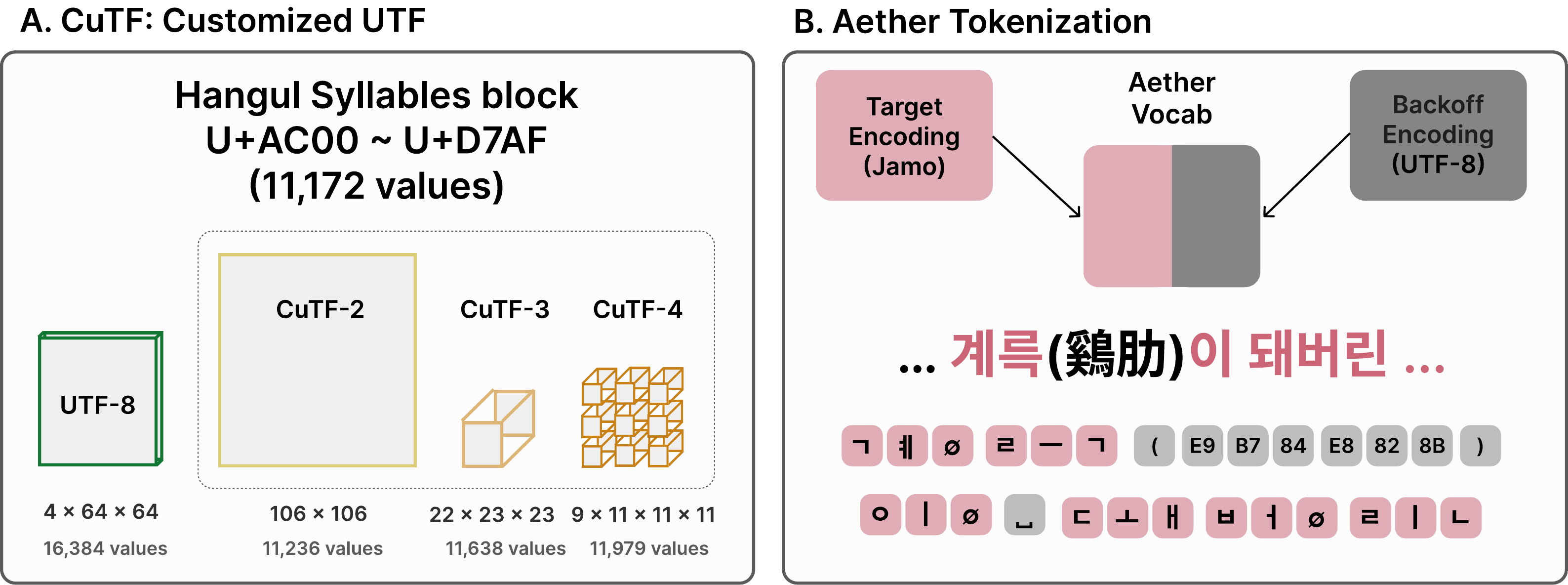}
    \caption{\textbf{A.} CuTF is a custom encoding scheme that maps target characters to a minimal vocabulary at a given compression rate. For example, \kr{강} would be encoded as [EA, B0, 95] out of [4, 64, 64] possible inventory elements in UTF-8, versus [46, 23, 21] out of position specific inventories of [22$_i$, 23$_j$, 23$_k$] in CuTF-3. \textbf{B.} AetherTokenizers use a hybrid representation that applies a specialized encoding to the target script and UTF-8 to all other text.}
    \label{fig:method}
\end{figure*}
\subsubsection{Orthography-aware Representations}

Drawing on prior work, we explore representation methods that exploit the orthographic structure of three scripts. To our knowledge, there are no script-specific encodings for Georgian.

\paragraph{Chinese}

The Chinese Hanzi script (mainly CJK Ideographs in Unicode) contain rich features, such as phonetic information and glyph information~\citep{wang2020radical, si-etal-2023-sub,xiang-liu-2025-exploiting}.
However, methods based on these features are often not suitable to encode  characters, as they require explicit symbols for character boundaries or disambiguation.
We experiment with character representations utilizing Ideographic Description Sequence (IDS) decomposition, which represents a character by its subcharacters and their arrangement~\citep{han-etal-2021-chinese}.
For instance, the character 悬~is decomposed to (⿱, 县, 心).
IDS sequences are self-delimiting and rarely require disambiguation, making them a viable method for character representation.
The IDS subcharacters are often also decomposable, allowing for recursive decomposition.
In the previous decomposition, subcharacter 心~has no further decompositions, but 县~can be further decomposed to (⿱, 且, and 厶).
Allowing one decomposition step (\textbf{IDS-1}) reduces the vocabulary size from 20,902 to 3,006 and allowing two steps of decomposition (\textbf{IDS-2}) reduces it further to 1,138.
Characters with identical decompositions are rare (20 in IDS-1 and 24 in IDS-2), and are addressed by a single additional disambiguation symbol.
This representation benefits from subcharacter information transferring across multiple characters, but may suffer from long and irregular encoding lengths.
We use the \texttt{cjkvi-ids} database\footnote{\url{https://github.com/cjkvi/cjkvi-ids}} for IDS decompositions.

\paragraph{Korean}
The Korean Hangul script (Hangul Syllables) have a compositional structure, where each character can be canonically decomposed into three sub-character alphabet units called \textit{jamo}~\citep{stratos-2017-sub,moon-okazaki-2020-jamo, cognetta-etal-2023-parameter}.
\kr{
For instance, the character 강~(\textit{gang}) can be losslessly decomposed into jamos that carry phonetic information: ㄱ~(\textit{g}), ㅏ~(\textit{a}), ㅇ~(\textit{-ng}).
Jamo decomposition of Hangul syllables can be expressed as $(c_h,c_v,c_t)$ with $c_h \in J_{h}$, $c_v\in J_v$, and $c_t\in J_t$ where $|J_{h}|=19, |J_{v}|=21, |J_{t}|=28$.
We test two variants of jamo-based encoding schemes.
The first, \textbf{Jamo} (fixed-length), always decomposes characters into three jamos (e.g. 하 $\rightarrow$ (ㅎ, ㅏ, $\emptyset$)), while the second \textbf{Jamo-varlen} encodes into two jamos whenever possible (e.g. 하 $\rightarrow$ (ㅎ, ㅏ))\footnote{The null character does not exist in this variant, therefore $|J_{t}|=27$.}.
The former may benefit from the regularity of a fixed length encoding, while the latter may benefit from stronger compression.
}

\paragraph{Tamil}
For abugida writing systems such as Tamil, a single user-perceived character (grapheme) may be encoded as multiple Unicode characters~\citep{velayuthan2024egalitarianlanguagerepresentationlanguage,nisfer2026graphemekitgraphemelevelmetricstext}.
For instance, the word \ta{வீடு} (\textit{veedu}, `house') is rendered as two graphemes \ta{வீ} (\textit{vee}) and \ta{டு} (\textit{du}), but is stored as four Unicode characters (\ta{'வ', 'ீ', 'ட', 'ு'}).
Following this prior work, we represent Tamil text at the \textbf{Graphemes} granularity, which better aligns with the units perceived by Tamil readers.
We follow the implementation of \citet{nisfer2026graphemekitgraphemelevelmetricstext}, which defines a total of 314 total graphemes, including 247 standard Tamil letters, 65 Grantha forms, and 2 conjunct graphemes (\ta{ஶ்ரீ, க்ஷ}).

\subsection{CuTF: Customized UTFs}
\label{sec:enc-cutf}
Generally, existing encodings are designed with the constraint of using computational units (i.e. bytes) as code units.
Using these encodings for language modeling unnecessarily inherits these constraints, leading to inefficient indexing.
We propose CuTF (Customized UTF; Figure~\ref{fig:method}A), a novel indexing scheme that encodes a target Unicode block using the minimum base units needed to index all codepoints in a target Unicode block at fixed lengths, relaxing the assumption that the total vocabulary of the encoding must fit within one byte (256 values).

Each CuTF implementation is parameterized by an encoding length $L$ (denoted \textbf{CuTF-L}), the number of units needed to represent a character in the target Unicode block. Specifically, the character is represented as a fixed-length
sequence of $L$ units $[c_1, \ldots, c_L]$, where the $i$-th unit is drawn
from a position-specific vocabulary $C_i$. Indexing within this encoding uses the offset within the target block rather than the Unicode codepoint.\footnote{For instance, instead of indexing `\kr{강}'~as the 44,053rd character in Unicode, CuTF indexes it as the 21st character in the Hangul Syllables block.}

The minimum vocabulary needed to fully map a block of size $V$ with length $L$ is $\lceil V^{1/L} \rceil$ per encoding unit. 
For instance, the 11,172 characters in the Hangul Syllables blocks can be mapped with three encoding units if each unit has 23 unique values.
In our experiments, we use $L=2,3,4$ for larger script inventories (\texttt{ZH}, \texttt{KO}) and $L=2,3$ for smaller scripts (\texttt{TA}, \texttt{KA}).

\section{Methodology}
We investigate the effectiveness of alternative encoding strategies (Section~\ref{sec:alt-encodings}) for monolingual language modeling by training comparable language models that vary only in their encoding method for the target script. We train each language's models on identical monolingual corpora for one epoch, introducing hybrid encoding schemes and character-level macro-averaged metrics to prevent out-of-vocabulary confounds and frequency biases in our evaluations, respectively.


\paragraph{Aether Tokenization}
Even in monolingual text corpora, characters outside the primary script (e.g. foreign words, symbols, emojis) can naturally occur. However,
many language-specific encodings do not provide full coverage of Unicode characters, leading to inconsistent handling of out-of-script characters across the considered settings. 
To in order to include the language-specific encodings while not introducing confounds, we propose \textsc{Aether} tokenizers, which apply a script-specific encoding only to the target script, while using UTF-8 as a back-off encoding for other characters (Fig~\ref{fig:method}B); this hybrid encoding set up enables fair comparison between encoding methods in their ability to represent a target script, without out-of-vocabulary issues.
A preprocessing regex separates target from non-target text to facilitate the selective application of encodings across inputs.
The target and fallback encoding spaces are stored in separate inventories, preventing cross-script collisions in model inputs.

\paragraph{Model Training}
For each encoding scheme, we train GPT2-base-sized decoder-only models with Llama-3 architecture from scratch, increasing sequence length from 512 to 4096 to account for longer sequences from finer-grained tokenization.
We use a maximum learning rate of 5e-4 at batch sizes of 64 using BF16 mixed precision training.
Additional hyperparameter information can be found in Appendix~\ref{appendix-training}.

Importantly, we train each model on one epoch, so that all models of the same language are trained on the same amount of data regardless of input sequence lengths.
Ensuring fair comparisons between models with different tokenizers is difficult, since runs cannot be both compute-matched~\citep{liu2025superbpe} and data-matched~\citep{park-etal-2020-empirical} at the same time.
We focus on data-matching over compute-matching in our experiments as it controls for the overall information seen by each model, while compute-matching benefits more compressive encodings artificially by allowing it to train on significantly more data. 

Each model is trained on a high-quality, curated pretraining text corpus for the target language.
For Chinese, we use FineWeb-Edu-Chinese-v2 from the OpenCSG Corpus~\citep{yu2025opencsgchinesecorpusseries}.
For Korean, we use the LLM pretraining corpus from the AI Hub\footnote{\url{https://www.aihub.or.kr/aihubdata/data/view.do?dataSetSn=71748}}.
For Tamil, we use the Tamil subset from the Sangraha dataset~\citep{khan2024indicllmsuite}.
For Georgian, we use the Georgian subset from the FineWeb2 dataset~\citep{penedo2025fineweb2pipelinescale}.
Each training corpus is capped to approximately 6 billion characters, corresponding to around 15 billion UTF-8 bytes for each language.
For comparison, this exceeds the 7--11 billion UTF-8 byte range (varying by language) recommended for our model size by tokenization scaling laws~\citep{limisiewicz2026computeoptimaltokenization}. 

For each setting, we report mean and standard deviation from 5 random seeds.
\paragraph{Evaluation}
For evaluation, we use language modeling probabilities of sequences taken from the FLORES dataset~\citep{goyal-etal-2022-flores}.
Our aim is to evaluate an encoding's ability to effectively represent all targeted characters in a script.
However, common characters are more frequent than uncommon characters by large magnitudes~\citep{Choi01042000, liu2017characterdistributionsclassicalchinese}, skewing standard language modeling metrics to a small subset of characters.
Therefore, we propose character-level macro averaged negative log-likelihood (\textbf{Macro NLL}) as our primary metric. Individual token NLLs are aggregated to the character-level\footnote{For Tamil, it is aggregated to the grapheme-level.} and averaged across character occurrences; then, the log-likelihood values of all unique characters are averaged together with equal weighting regardless of frequency.
For experimental results with the standard bits-per-character loss, refer to Appendix~\ref{appendix-bpc}.

\begin{figure*}[t]
\begin{center}
\includegraphics[width=0.9\linewidth]{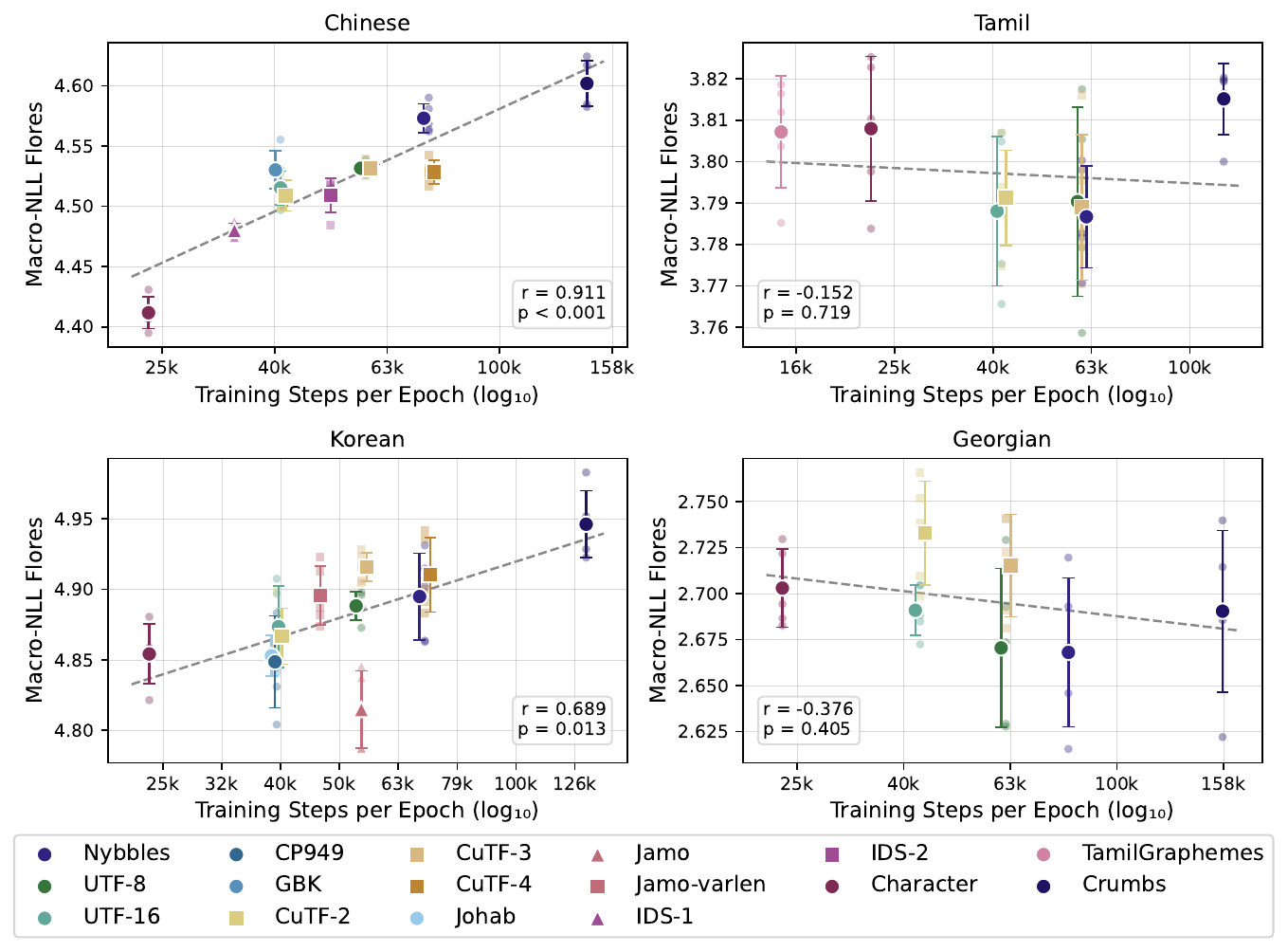}
\end{center}
\caption{Macro negative log-likelihoods (lower is better) of alternative text encodings on all considered languages. A linear regression model shows the relationship between Macro-NLL and log(training steps per epoch). X-values are jittered for visualization.} 
\label{results-linearfits}
\end{figure*}


\section{Results}

Figure~\ref{results-linearfits} shows log-linear fits between the compression of an encoding and its language modeling performance.
Since all models for a given language are trained on the same data for one epoch, we measure compression as the number of training steps per epoch.
We observe two distinct trends between languages.
\paragraph{Chinese and Korean}
Encoding compression was significantly correlated with language modeling performance, with more compressive encodings performing better.
The correlation was strong for Chinese ($r = 0.911$, $p < 0.001$) and moderately strong for Korean ($r = 0.689$, $p = 0.013$).
We note that the weaker correlation for Korean was largely driven by the Jamo encoding.
Excluding it yields a correlation comparable to Chinese ($r = 0.886$, $p < 0.001$).

\paragraph{Tamil and Georgian}
Encoding compression was not significantly correlated with language modeling performance for either Tamil ($r = -0.152$, $p = 0.719$) or Georgian ($r = -0.376$, $p = 0.405$).
Overlapping error bars suggest that variability between encodings is comparable to variability across random seeds.
Even in the most extreme comparisons, encodings performed similarly despite large differences in compression: Graphemes and Crumbs differed by a factor of ${\sim}8$ for Tamil, as did Character and Crumbs by a factor of ${\sim}7$ for Georgian.
Although neither correlations is significant, they trend in the opposite direction from Chinese and Korean.

\section{Discussion} \label{sec:discussion}

\begin{figure}[t]
\begin{center}
\includegraphics[width=0.9\columnwidth]{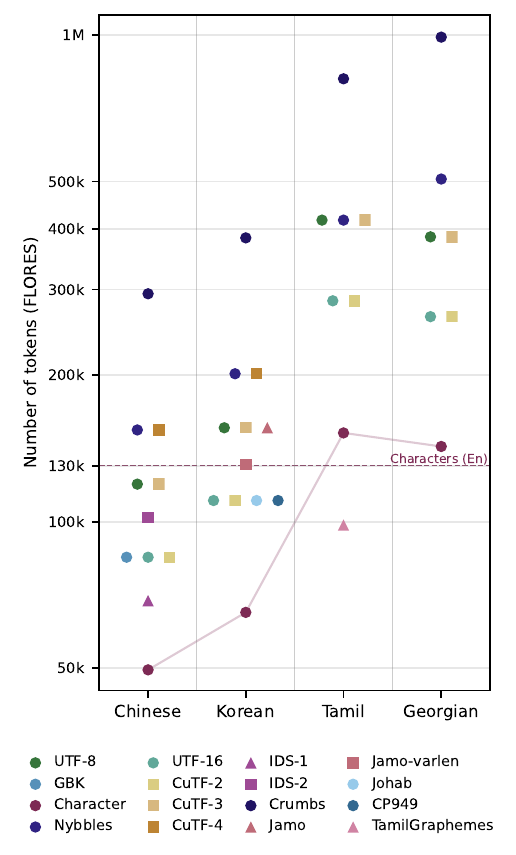}
\end{center}
\vspace{-0.4cm}
\caption{Number of tokens produced by different encodings on parallel data (\mbox{FLORES}). Dotted red line marks the character counts in English for comparison.}
\label{discussion-charcomps}
\end{figure}

\paragraph{Encoding Effects are Language-specific.}
Why does encoding choice impact Chinese and Korean but not (significantly) Tamil and Georgian?
We attribute this difference to \textit{orthographic granularity}, or how densely languages represent linguistic content in writing~\citep{shani2026rootsperformancedisparitymultilingual}.
In Chinese, each character typically corresponds to a morpheme, and in Korean, each character encodes a full syllable.
In contrast, characters in Tamil and Georgian correspond to individual consonants or vowels, which contain little information in isolation.
Consistent with this, Chinese and Korean require less than half as many characters as Tamil and Georgian to express the same content in our evaluation data (Figure~\ref{discussion-charcomps}).
We hypothesize that when characters contain substantial information, allowing models to reconstruct this information in fewer steps (fewer tokens) is beneficial for performance.
Conversely, when characters carry negligible information on their own, characters and subcharacters perform similarly.

\paragraph{Role of Compression.}
Tokenizer compression has been said to be correlated with model performance~\cite{galle-2019-investigating, goldman2024unpackingtokenizationevaluatingtext}, but there is conflicting evidence~\cite{schmidt-etal-2024-tokenization}.
It may be the case that longer sequences from finer-grained tokenization are advantageous, as this allows models to spend more FLOPs on a given text~\cite{xue-etal-2022-byt5, marathe2026breaking}.
\citet{limisiewicz2026computeoptimaltokenization} find increasing compression past an optimal point can harm performance.
Our findings suggest that the role of compression can be heavily dependent on the language.

\paragraph{Linguistically Meaningful Encoding Units.}
For Korean models, the best performing Jamo encoding outperformed the most compressive encoding (Character).
Jamo encoding decomposes each syllable into three jamos, a meaningful sub-character unit that can capture linguistic phenomena such as morphophonological alternations~\citep{kim-etal-2026-script}.
In contrast, Jamo-varlen, which uses the same sub-character units but with variable-length sequences, did not show the same benefits and instead performed in line with its compression.
Similarly, IDS-1 and IDS-2 also performed in line with their compression, despite using linguistically informative subcharacter units~\citep{han-etal-2021-chinese} as variable-length sequences.
This suggests that sub-character information is only useful in settings where it is paired with fixed-length encoding patterns.

\paragraph{What's in a good encoding?}
Recent work argues that tokenization should be considered as a core design decision appropriately tailored for its context~\citep{alqahtani-etal-2026-stop}.
Encodings, much like subword tokenizers, have desiderata that should be considered before being deployed in the language modeling pipeline.
Even without performance benefits, compression is a desirable quality for efficient model training and inference.
However, higher compression often necessitates a larger vocabulary size (as seen in Table~\ref{tab:encodings}).
Our findings on when and how compression affects performance can help inform these tradeoffs when choosing an encoding.

Another factor is encoding fairness between scripts.
Our approach of using hybrid encodings of Aether tokenization poses a solution to encoding length disparities in UTF-8~\citep{land2025bpestaysscriptstructured,arnett-etal-2024-bit}.
Hybrid encodings provide a method to improve compression or performance at the cost of increased vocabulary size.


\paragraph{Do these results extend to subword tokenizers?}
We argue that understanding the effect of encoding choice is relevant for all tokenizers.
All modern subword tokenization models (byte-level BPE, and even character-level BPE or Unigram with byte fallback) implicitly use UTF-8 encoding as a pretokenizer, which has tangible and underlooked effects on the final tokenization~\citep{velayuthan2024egalitarianlanguagerepresentationlanguage,gorman-pinter-2025-dont}.
Our work suggests that the default UTF-8 encoding used by subword tokenizers may disadvantage certain scripts with longer byte lengths, since specialized encodings can provide increased compression and improved language modeling performance. This effect might be the most significant for languages with coarser orthographic granularities like Korean and Chinese.

\section{Conclusion}
In this work, we investigate alternative encoding schemes for their impact on language modeling performance.
We conduct a comprehensive analysis of encoding effects on 38 language-encoding settings, considering Unicode-based encodings, language-specific encodings, and our novel encoding scheme, CuTF, that efficiently encodes the target script at a chosen compression rate.
To facilitate full coverage of possible characters in language-specific encoding experiments, we also introduce Aether Tokenization, a new hybrid encoding method that can reap the benefits of specialized encodings without sacrificing character coverage.
We find that byte-level language models benefit from compressive text encodings for languages with coarse orthographic granularity (such as Chinese and Korean), while languages with finer granularity (such as Tamil and Georgian) are largely insensitive to encoding choice.
In light of these results, we argue that practitioners should not default to UTF-8 when building a language model, but instead treat the choice of encoding as a deliberate design decision that considers the targeted languages.



\section{Limitations}

Given the computational costs of comprehensively analyzing multiple encoding schemes across languages and modeling runs, the models considered in this work are relatively small, at approximately 120M parameters. Scaling these experiments in future work will be important for verifying our results, particularly on downstream tasks beyond next-word prediction. Additionally, our evaluation is currently limited to monolingual settings, though we hypothesize that encoding effects may behave differently in multilingual settings due to cross-script interactions and the availability of alternative encodings for different languages. 


Finally, like standard byte-level language models using UTF-8, models trained on our custom encodings can generate invalid or malformed sequences \citep{moon2026perplexityutf8validitybyteaware, firestone2025utf}, an issue we leave unaddressed in this paper. We anticipate that some encoding schemes may be particuarly prone to this issue, such as IDS tokenization, that would require the model to precisely match long, variable-length sequences for each character when generating new text.

\bibliography{custom}

\appendix
\section{Training Details}
\label{appendix-training}
The hyperparameters of our models are found in Table~\ref{tab:hyperparameters}.
For hyperparameter search, we performed a sweep over {1e-4, 5e-4, 1e-3} learning rate on three encodings (UTF-8, Jamo, and GBK) on Chinese and Korean.
Higher learning rate values would produce better training curves, but were frequently prone to training loss spikes.
The 5e-4 value was chosen for our experiments, but occasional training loss spikes persisted.
Increasing batch size (64$\rightarrow$128) improved stability but harmed performance.
Eventually, changing $\beta_2$ value of the AdamW optimizer from the default 0.999 value to 0.95 improved both stability and performance.

\begin{table}[th]
    \centering
    
      \begin{tabular}{lc}
        \toprule
        Hyperparameter & Value\\
        \midrule
        Number of Layers & 12  \\
        Hidden Size & 768 \\
        Inner Hidden Size & 3072 \\
        Attention Heads & 12 \\
        Attention Dropout & 0.0 \\
        Max Sequence Length & 4096 \\
        Peak Learning Rate & 5e-4 \\
        Batch Size & 64 \\
        Weight Decay & 0.01 \\
        Learning Rate Decay & Linear \\
        LR Warmup & 0.1 \\
        Adam $\epsilon$ & 1e-8 \\
        Adam $\beta_{1}$ & 0.9 \\
        Adam $\beta_{2}$ & 0.95 \\
        Max Grad Norm & 1 \\
      \bottomrule
    \end{tabular}%
    \caption{The hyperparameters used for training our models.}
    \label{tab:hyperparameters}
\end{table}
\onecolumn

\section{BPC Results}
\label{appendix-bpc}
\begin{figure*}[h]
\begin{center}
\includegraphics[width=0.9\linewidth]{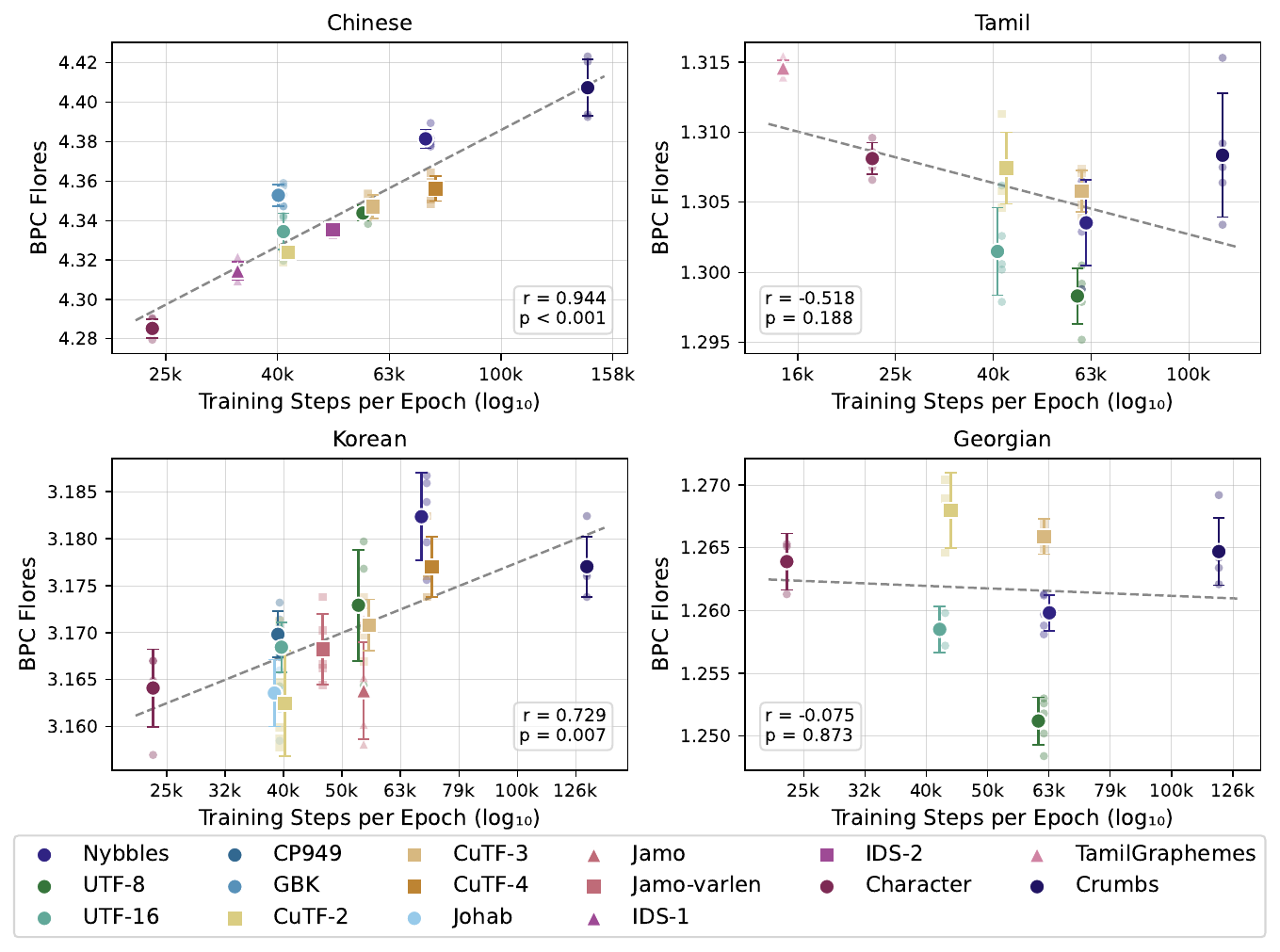}
\end{center}
\caption{Bits-per-character (lower is better) of alternate text encodings for the considered languages. A linear regression model shows the relationship between BPC and log(training steps per epoch). X-values are jittered for visualization.}
\label{results-micro}
\end{figure*}

Figure~\ref{results-micro} contains our results using the BPC (bits-per-character) metric.
Overall trends remain the same, but the impact of long-tailed characters are diminished in this metric.

\end{document}